\documentclass[10pt,twocolumn,letterpaper]{article}

\usepackage[pagenumbers]{cvpr}              %

\usepackage[accsupp]{axessibility}

\makeatletter
\@namedef{ver@everyshi.sty}{}
\makeatother

\usepackage{acronym}
\usepackage{adjustbox}
\usepackage{booktabs}
\usepackage[font=small,labelfont=bf]{caption}
\usepackage{colortbl}
\usepackage{fmtcount}
\usepackage{forloop}
\usepackage{multirow}
\usepackage{stackengine}
\usepackage{suffix}
\usepackage[para]{footmisc}  
\usepackage{svg}
\usepackage{rotating}
\usepackage{pifont}    
\usepackage{stmaryrd}  
\usepackage{tabularx}
\usepackage[textsize=scriptsize]{todonotes}
\usepackage{xfrac}
\usepackage{xr}
\usepackage{xspace}

\usepackage{ragged2e}
\usepackage{microtype}
\usepackage{setspace}
\usepackage[scaled=0.95]{inconsolata}

\presetkeys{todonotes}{inline}{}  

\makeatletter
\newcommand*{\addFileDependency}[1]{
  \typeout{(#1)}
  \@addtofilelist{#1}
  \IfFileExists{#1}{}{\typeout{No file #1.}}
}
\makeatother

\definecolor{ForestGreen}{HTML}{228b22}  

\DeclareRobustCommand{\nbd}{\nobreakdash-} 

\newcommand{\down}{\textcolor{black}{\ensuremath{\boldsymbol{\downarrow}}}}
\newcommand{\up}{\textcolor{black}{\ensuremath{\boldsymbol{\uparrow}}}}

\newcommand{\heading}[1]{\noindent \textbf{\small #1.}}

\definecolor{Asectioncolor}{RGB}{255, 200, 200}
\definecolor{Bsectioncolor}{RGB}{255, 228, 196}
\definecolor{Csectioncolor}{RGB}{235, 255, 235}
\definecolor{Dsectioncolor}{RGB}{235, 235, 255}

\newcommand{\addfig}[3][htbp]{%
\begin{figure}[#1]

\centering
\input{Figures/#3}
\label{fig:#3}

\end{figure}
}

\WithSuffix\newcommand\addfig*[3][htbp]{%
\begin{figure*}[#1]

\centering
\input{Figures/#3}
\label{fig:#3}

\end{figure*}
}

\newcommand{\addtbl}[3][htbp]{%
\begin{table}[#1]

\scriptsize
\addtolength{\tabcolsep}{-0.2em}
\renewcommand{\arraystretch}{1.1}
\centering
\input{Tables/#3}
\label{tbl:#3}

\end{table}
}

\WithSuffix\newcommand\addtbl*[3][htbp]{%
\begin{table*}[#1]

\scriptsize
\addtolength{\tabcolsep}{-0.2em}
\renewcommand{\arraystretch}{1.1}
\centering
\input{Tables/#3}
\label{tbl:#3}

\end{table*}
}

\newcommand{\fig}[1]{Figure~\ref{fig:#1}}
\newcommand{\tbl}[1]{Table~\ref{tbl:#1}}

\newcommand{\acromath}[3]{\acrodef{#1}[\(#2\)]{#3}} 
\newcommand{\myac}[1]{\text{\acs{#1}}}  

\newcommand{\shape}[4]{\ensuremath{  
#1 \times #2
\ifthenelse {\equal{#3}{}} {} {\times #3}
\ifthenelse {\equal{#4}{}} {} {\times #4}
}}

\newcommand{\sca}[1]{\ensuremath{#1}}  						   
\newcommand{\vct}[1]{\ensuremath{\textbf{\MakeLowercase{#1}}}} 
\newcommand{\mat}[1]{\ensuremath{\textbf{\MakeUppercase{#1}}}} 
\newcommand{\set}[1]{\ensuremath{\MakeUppercase{#1}}} 	       

\newcommand{\thr}{\delta}

\definecolor{firstcolor}{HTML}{BDE6CD}
\definecolor{secondcolor}{HTML}{E2EEBC}
\definecolor{thirdcolor}{HTML}{FFF8C5}

\acrodef{ar}     [AR]             {Augmented Reality}
\acrodef{araug}  [AR-Aug]         {aspect ratio augmentation}

\acrodef{cnn}    [CNN]            {Convolutional Neural Network}

\acrodef{d25}    [$\delta_{.25}$] {$\delta < 1.25$}
\acrodef{dnn}    [DNN]            {Deep Neural Network}

\acrodef{fscore} [F]              {F-Score}

\acrodef{imu}    [IMU]            {Inertial Measurement Unit}

\acrodef{kbr}    [KBR]            {Kick Back \& Relax}

\acrodef{lidar}  [LiDAR]          {Light Detection and Ranging}

\acrodef{mae}    [MAE]            {Mean Absolute Error}
\acrodef{mde}    [MDE]            {monocular depth estimation}
\acrodef{mdec}   [MDEC]           {Monocular Depth Estimation Challenge}
\acrodef{ml}     [ML]             {Machine Learning}

\acrodef{rel}    [Rel]            {Absolute Relative Error}

\acrodef{sfm}    [SfM]            {Structure-from-Motion}
\acrodef{slam}   [SLAM]           {Simultaneous Localization and Mapping}
\acrodef{sota}   [SotA]           {State-of-the-Art}
\acrodef{ssl}    [SSL]            {self-supervised learning}

\acrodef{vo}     [VO]             {Visual Odometry}

\acrodef{crf}      [NeWCRFs] {NeWCRFs}

\acrodef{ddad}     [DDAD]    {DDAD}
\acrodef{dii}      [DIODE]   {DIODE Indoors}
\acrodef{dio}      [DIODE]   {DIODE Outdoors}
\acrodef{dpt-beit} [DPT]     {DPT-BEiT}
\acrodef{dpt-vit}  [DPT]     {DPT-ViT}

\acrodef{kb}       [KB]      {Kitti Benchmark}
\acrodef{ke}       [KE]      {Kitti Eigen}
\acrodef{keb}      [KEB]     {Kitti Eigen{\nbd}Benchmark}
\acrodef{kez}      [KEZ]     {Kitti Eigen{\nbd}Zhou}

\acrodef{mc}       [MC]      {Mannequin Challenge}
\acrodef{midas}    [MiDaS]   {MiDaS}

\acrodef{nyud}     [NYUD]    {NYUD{\nbd}v2}

\acrodef{sintel}   [Sintel]  {Sintel}
\acrodef{ssmde}    [SS-MDE]  {SS{\nbd}MDE}
\acrodef{stv}      [STV]     {SlowTV}
\acrodef{syns}     [SYNS]    {SYNS{\nbd}Patches}

\acrodef{tum}      [TUM]     {TUM{\nbd}RGBD}

\acromath{pix}        { \vct{p} } {}

\acromath{point-gt}   { \vct{q} } {}
\acromath{point-pred} { \hat{\myac{point-gt}} } {}
 
\acromath{depth-gt}   { \sca{y} } {}
\acromath{depth-pred} { \hat{\myac{depth-gt}} } {}
 
\acromath{cloud-gt}   { \set{Q} } {}
\acromath{cloud-pred} { \hat{\myac{cloud-gt}} } {}
 
\acromath{edges-gt}   { \mat{M} } {}
\acromath{edges-pred} { \hat{\acs{edges-gt}} } {}
 
\acromath{thr-3d}     { \sca{\thr} } {}

\definecolor{cvprblue}{rgb}{0.21,0.49,0.74}
\usepackage[breaklinks,colorlinks,citecolor=cvprblue]{hyperref}

\graphicspath{{Images/}}
\def\codelink{\href{https://github.com/toshas/mdec_benchmark}{https://github.com/toshas/mdec\_benchmark}}
\def\challink{\href{https://codalab.lisn.upsaclay.fr/competitions/21305}{https://codalab.lisn.upsaclay.fr/competitions/21305}}

\newcommand{\team}[1]{%
\ifcase#1
\textbf{Baseline}\xspace
\or HRI\xspace
\or Lavreniuk\xspace
\or Mach-Calib\xspace
\or EasyMono\xspace
\or Insta360-Percep\xspace
\or  HIT-AIIA\xspace
\or Robot02-vRobotit\xspace
\or ViGIR LAB\xspace
\or HCMUS-DepthFusion\xspace
\or ReadingLS\xspace
\else
Invalid number%
\fi
}

\newcommand{\train}[1]{%
\ifcase#1
S\xspace
\or $\dagger$D*\xspace
\or $\dagger$D*\xspace
\or $\dagger$D*\xspace
\or $\dagger$D\xspace
\or $\dagger$D\xspace
\or $\dagger$D\xspace
\or $\dagger$D\xspace
\or $\dagger$D\xspace
\or $\dagger$D\xspace
\or $\dagger$D\xspace
\or $\dagger$D\xspace
\or ?\xspace
\or ?\xspace
\or ?\xspace
\or D\xspace
\or $\dagger$MD*\xspace
\or ?\xspace
\or ?\xspace
\or D\xspace
\else
Invalid number%
\fi
}

\newcommand{\teamsec}[1]{Team #1: \team{#1} } %

\usepackage[capitalize]{cleveref}
\crefname{section}{Sec.}{Secs.}
\Crefname{section}{Section}{Sections}
\Crefname{table}{Table}{Tables}
\crefname{table}{Tab.}{Tabs.}

\def\confName{CVPR}
\def\confYear{2024}

\makeatletter
\let\@fnsymbol\@arabic
\makeatother

\begin{document}

\title{The Fourth Monocular Depth Estimation Challenge}

\author{%
\begin{minipage}{0.9\textwidth}%
\noindent%
\justifying%
\setstretch{1.25}%
Anton Obukhov$^{\dagger 1}$ \hfill
Matteo Poggi$^{\dagger 2}$ \hfill
Fabio Tosi$^{\dagger 2}$ \hfill
Ripudaman Singh Arora$^{\dagger 3}$ \\
Jaime Spencer$^{\dagger 4}$ \hfill
Chris Russell$^{\dagger 5}$ \hfill
Simon Hadfield$^{\dagger 6}$ \hfill
Richard Bowden$^{\dagger 6}$ \\[0.2cm]
Shuaihang Wang$^{7}$ \hfill
Zhenxin Ma$^{7}$ \hfill
Weijie Chen$^{7}$ \hfill
Baobei Xu$^{7}$ \hfill
Fengyu Sun$^{7}$ \hfill
Di Xie$^{7}$ \\
Jiang Zhu$^{7}$ \hfill
Mykola Lavreniuk$^{8}$ \hfill
Haining Guan$^{9}$ \hfill
Qun Wu$^{9}$ \hfill
Yupei Zeng$^{9}$ \hfill
Chao Lu$^{9}$ \\
Huanran Wang$^{9}$ \hfill
Guangyuan Zhou$^{4}$ \hfill
Haotian Zhang$^{4}$ \hfill
Jianxiong Wang$^{4}$ \hfill
Qiang Rao$^{4}$ \\
Chunjie Wang$^{4}$ \hfill
Xiao Liu$^{4}$ \hfill
Zhiqiang Lou$^{10}$ \hfill
Hualie Jiang$^{10}$ \hfill
Yihao Chen$^{10}$ \hfill
Rui Xu$^{10}$ \\
Minglang Tan$^{10}$ \hfill
Zihan Qin$^{11}$ \hfill
Yifan Mao$^{11}$ \hfill
Jiayang Liu$^{11}$ \hfill
Jialei Xu$^{11}$ \hfill
Yifan Yang$^{11}$ \\
Wenbo Zhao$^{11}$ \hfill
Junjun Jiang$^{11}$ \hfill
Xianming Liu$^{11}$ \hfill
Mingshuai Zhao$^{12}$ \hfill
Anlong Ming$^{12}$ \\
Wu Chen$^{12}$ \hfill
Feng Xue$^{13}$ \hfill
Mengying Yu$^{12}$ \hfill
Shida Gao$^{12}$ \hfill
Xiangfeng Wang$^{12}$ \\
Gbenga Omotara$^{14}$ \hfill
Ramy Farag$^{14}$ \hfill
Jacket Demby’s$^{14}$ \hfill
Seyed Mohamad Ali Tousi$^{14}$ \\
Guilherme N. DeSouza$^{14}$ \hfill
Tuan-Anh Yang$^{15,16}$ \hfill
Minh-Quang Nguyen$^{15,16}$ \\
\mbox{\hspace{3em}}
Thien-Phuc Tran$^{15,16}$ \hfill
Albert Luginov$^{17}$ \hfill
Muhammad Shahzad$^{17}$
\mbox{\hspace{3em}}
\end{minipage}
}

\maketitle

\begin{abstract}
This paper presents the results of the fourth edition of the Monocular Depth Estimation Challenge (MDEC), which focuses on zero-shot generalization to the SYNS-Patches benchmark, a dataset featuring challenging environments in both natural and indoor settings.  
In this edition, we revised the evaluation protocol to use least-squares alignment with two degrees of freedom to support disparity and affine-invariant predictions.  
We also revised the baselines and included popular off-the-shelf methods: Depth Anything v2 and Marigold.
The challenge received a total of 24 submissions that outperformed the baselines on the test set; 10 of these included a report describing their approach, with most leading methods relying on affine-invariant predictions.  
The challenge winners improved the 3D F-Score over the previous edition’s best result, raising it from 22.58\% to 23.05\%.  
\end{abstract}

\makeatletter
\let\orig@makefnmark\@makefnmark
\let\orig@makefntext\@makefntext
\let\@makefnmark\relax
\renewcommand{\@makefntext}[1]{%
  \noindent{\@makefnmark}#1%
}
\makeatother
\footnotetext{
\hangindent=0em
\hangafter=0
\noindent
$\dagger$ Challenge Organizers \\[5pt]
$^1$Huawei Research Zürich \hspace{1em}
$^2$University of Bologna \hspace{1em}
$^3$Blue River Technology \hspace{1em}
$^4$Independent Researcher \hspace{1em}
$^5$Oxford Internet Institute \hspace{1em}
$^6$University of Surrey \hspace{1em}
$^7$Hikvision Research Institute \hspace{1em}
$^8$Space Research Institute NASU-SSAU, Kyiv, Ukraine \hspace{1em}
$^9$Megvii \hspace{1em}
$^{10}$Insta360 \hspace{1em}
$^{11}$Harbin Institute of Technology \hspace{1em}
$^{12}$Beijing University of Posts and Telecommunications \hspace{1em}
$^{13}$University of Trento \hspace{1em}
$^{14}$University of Missouri \hspace{1em}
$^{15}$University of Science, VNU-HCM, Vietnam \hspace{1em}
$^{16}$Vietnam National University, Ho Chi Minh, Vietnam \hspace{1em}
$^{17}$University of Reading
}

\makeatletter
\let\@makefnmark\orig@makefnmark
\let\@makefntext\orig@makefntext
\makeatother

\section{{Introduction}} \label{sec:intro}
Monocular depth estimation (MDE) is an important low-level vision task with applications in fields such as augmented reality, robotics, and autonomous vehicles. 
In 2024, the field was dominated by generative approaches, with DepthAnything~\cite{Yang2024,Yang2024v2} representing the transformer-based solution and Marigold~\cite{ke2024repurposing} being a denoising diffusion model based on the popular Text-to-Image LDM~\cite{rombach2022high}.

The MDE task aims to recover a notion of depth for each pixel of a single input image. 
This could take, for example, the form of a metric depth, where each value is a metric distance from the corresponding visible object in the scene to the camera capturing it.
More often than that, disparity or inverse depth encodes reciprocal values of the metric depth up to some unknown scale, which is closely related to stereo vision systems.
Another universal representation is affine-invariant depth, in which each value represents a linear interpolation between the scene’s near and far planes.

The absence of multi-view geometric cues within a single image makes MDE highly ill-posed and, consequently, one of the longest-standing challenges in computer vision. Nevertheless, the emergence of deep learning has imbued this research with learned visual cues, enabling the development of increasingly accurate solutions year after year.

\addfig*[!t]{syns}{data:syns}

Early attempts to tackle this task were confined within single, limited domains, such as indoor environments \cite{Silberman2012} or urban driving scenarios \cite{Geiger2013}, resulting in highly-specialized solutions that failed to generalize beyond their specific domain of expertise.
To address this limitation, in the last three years, alongside to the development of more advanced frameworks and evaluation settings \cite{Ranftl2020,Ranftl2021}, the \ac{mdec} has pushed the research community toward more general models capable of cross-domain generalization, building upon a unique benchmark featuring a set of complex environments, comprising natural, agricultural, urban, and indoor settings. This benchmark includes a validation and a testing split, while not providing any training set -- thus necessitating inherent generalization capabilities from all methods.

Specifically, \acl{syns}~\cite{Adams2016,Spencer2022} was selected for this purpose, offering a diverse collection of environments, including urban, residential, industrial, agricultural, natural, and indoor scenes, all annotated with dense high-quality \acs{lidar} ground-truth, which is typically very challenging to acquire in outdoor settings.
This key characteristic, showcased in \fig{data:syns}, allows for a fair evaluation of MDE models, potentially free from some of the biases affecting popular benchmarks \cite{Geiger2013}.

The initial three editions of \ac{mdec}~\cite{Spencer_2024_CVPR} benchmarked self-supervised approaches first \cite{Spencer2023}, then expanded~\cite{spencer2023second,Spencer_2024_CVPR} to include supervised methods. All approaches were evaluated on both image-based and pointcloud-based metrics after the predicted depth maps were aligned with ground truth through median scaling \cite{Zhou2017}.
The fourth edition of \ac{mdec}, detailed in this paper, conducted in conjunction with CVPR'2025, refines the evaluation protocol used by the previous editions, primarily by introducing a robust least-squares alignment between predictions and ground-truth depth maps to adhere to the latest standards in the field \cite{Ranftl2020,Ranftl2021}. Additionally, given the recent advent of the first foundational models for depth estimation, we upgraded from the original baseline~\cite{Spencer2022,Garg2016} used in the first three editions with more contemporary models \cite{ke2024repurposing,Yang2024v2}, that better represent the latest advances in the field.

The availability of these models represents a strong starting point for any participant willing to compete in our challenge, which has already markedly increased overall participation in the third edition \cite{Spencer_2024_CVPR}. 
This trend continued this year, with 24 teams surpassing the \ac{sota} baseline in either pointcloud- or image-based metrics -- compared to 19 teams outperforming the old baseline last year. Among these competitors, 10 submitted a report detailing their approach; however, only a single team managed to outperform the winning team of the third edition.  
In the subsequent sections, we provide concise overviews of each submission, examine their performance on \acl{syns}, and explore promising future research directions.

\section{Related Work} \label{sec:lit}

\subsection{Single Domain MDE}
\heading{Supervised Methods}
The early advancements in monocular depth estimation primarily relied on single-domain supervised learning approaches with ground truth depth annotations. Eigen et al.~\cite{Eigen2015} introduced the first convolutional neural network for depth estimation, featuring a coarse-to-fine architecture and scale-invariant loss function. This inspired subsequent developments incorporating structured prediction models, including Conditional Random Fields (CRFs)~\cite{Liu2015,Weihao2022} and regression forests~\cite{Roy2016}. The field witnessed significant performance improvements through architectural innovations such as deeper networks~\cite{Xian2018,Ranftl2020}, multi-scale fusion techniques~\cite{Miangoleh2021}, and transformer-based encoders~\cite{Ranftl2021,Cheng2021,Bhat2023}. Some works reframed depth estimation as a classification problem~\cite{Fu2018,Li2019,Bhat2021,Bhat2022}, while others focused on developing novel loss functions, including gradient-based regression~\cite{Li2018,Wang2019b}, the berHu loss~\cite{Laina2016a}, ordinal relationship loss~\cite{Chen2016}, and affine-invariant mechanisms~\cite{Ranftl2020}.

\heading{Self-Supervised Methods}
Advances in supervised methods are bound to extensive and costly data collection and annotation campaigns. Years ago, the limited availability of labeled datasets ignited the development of self-supervised approaches. These techniques replace the supervision from depth labels with image-reconstruction losses, given the availability of a second image~\cite{Garg2016} of the same scene. This paradigm consists of exploiting image warping as supervision to the model and can be implemented either by using paired images collected through a calibrated stereo setup~\cite{Garg2016,Godard2017,Poggi2021}, or a single video collected by a moving camera~\cite{Zhou2017}.
Following these pioneering frameworks, researchers have pursued a number of directions to improve the accuracy of depth estimation. Some integrated feature-based reconstructions~\cite{Zhan2018,Spencer2020,Yu2020}, semantic segmentation~\cite{Ramirez2019}, adversarial losses~\cite{Aleotti2018}, and proxy-depth representations~\cite{Klodt2018,Rui2018,Andraghetti2019,Watson2019,Tosi2019,Choi2021,Peng2021}. Others explored trinocular supervision~\cite{Poggi2018} and additional geometric constraints~\cite{Mahjourian2018,Wang2018,Bian2019}. Significant effort was also directed toward improving depth estimates at object boundaries~\cite{Tosi2021,Talker2024}.
The presence of moving objects in videos represents a serious concern for self-supervised methods. Some approaches dealt with it by using uncertainty~\cite{Klodt2018,Yang2020a,Poggi2020}, predicting motion masks~\cite{Gordon2019,Casser2019,Dai2020,Tosi2020}, incorporating optical flow estimation~\cite{Yin2018,Ranjan2019,Luo2020}, or designing robust losses~\cite{Godard2019}. In a concurrent track, many works introduced improved architectural designs: on the one hand, to pursue higher accuracy~\cite{Guizilini2020,Bello2021,Zhao2022,Agarwal2023,Pillai2019,Lyu2021,Johnston2020,Yan2021,Zhou2021}, on the other hand, to deploy compact and efficient models~\cite{Poggi2018b,Peluso2019,Wofk2019,Aleotti2020,Peluso2021,Cipolletta2021,Huynh2022} that could bring self-supervised depth estimation closer to the deployment in real-time applications.

\subsection{Generalization and ``In-the-Wild'' MDE} 
Estimating depth in challenging, unconstrained environments demands methods that generalize to diverse, unknown settings~\cite{Chen2016,Chen2020}. While the previously outlined methods trained on single datasets struggled with generalization, newer approaches addressed this limitation by incorporating diverse data sources including large-scale datasets~\cite{Ranftl2021,Ranftl2020,Yang2024}, internet photo collections~\cite{Li2018,Yin2020}, automotive LiDAR~\cite{Geiger2013,Guizilini2020,Huang2019}, RGB-D sensors~\cite{Silberman2012,Cho2021,Sturm2012}, structure-from-motion~\cite{Li2018,Li2020}, optical flow~\cite{Xian2018,Ranftl2020}, and crowd-sourced annotations~\cite{Chen2016}. Among them, self-supervised approaches~\cite{Yin2021,Zhang2022} have also received increasing attention, with notable works like KBR(++)~\cite{Spencer2023c,Spencer2024} leveraging curated internet videos.

Furthermore, architectural improvements significantly advanced performance in this domain. The transition from CNNs to transformer-based models yielded substantial improvements, as demonstrated by DPT (MiDaS v3)~\cite{Ranftl2021} and Omnidata~\cite{Eftekhar2021}. A significant turning point has been the development of foundational models such as Depth Anything~\cite{Yang2024} and its successor Depth Anything v2~\cite{Yang2024v2}, which use large unlabeled data (\~62M images) to produce highly robust and generalizable depth estimators.

Another breakthrough emerged with generative approaches, particularly diffusion models~\cite{Song2020, Ho2020}. Early applications include Ji et al.~\cite{Ji2023}, Duan et al.~\cite{Duan2023}, and Saxena et al.~\cite{Saxena2023, Saxena2023b}. Afterwards, Marigold~\cite{ke2024repurposing} showed how diffusion-based generators can be adapted for MDE with rich prior knowledge. GeoWizard~\cite{Fu2024} extended an LDM~\cite{rombach2022high} to jointly predict depth and normal maps, enabling information exchange between these representations. \cite{zhang2024betterdepth} introduced a diffusion-based depth refiner, while \cite{viola2024marigolddc} explored guidance with sparse LiDAR inputs.
\cite{he2024lotus} and \cite{martingarcia2024diffusione2eft} developed simplified protocols for dense prediction.

Other approaches have focused on recovering metric depth by incorporating camera information, including ~\cite{Yin2023,hu2024metric3d,Guizilini2023,bochkovskii2024depth}. Meanwhile, works on monocular point map estimation ~\cite{Yin_2021_CVPR, yin2022towards, wang2024moge,piccinelli2024unidepth,piccinelli2025unidepthv2,wang2024dust3r} have explored direct 3D point recovery from single images.

Finally, video-based depth estimation has emerged as another direction, with approaches like \cite{shao2025chronodepth, hu2024depthcrafter, ke2024rollingdepth, chen2025video} addressing temporal consistency.

\subsection{Overcoming Challenging Conditions}
\heading{Adverse Weather} Monocular networks have historically struggled in adverse weather conditions. Various techniques have addressed these challenges, including methods for low visibility scenarios~\cite{Spencer2020}, day-night branches using GANs~\cite{Vankadari2020,Zhao2022b}, additional sensor integration~\cite{Gasperini2021}, and performance trade-offs~\cite{Vankadari2023}. A notable advance came with ~\cite{Gasperini2023}, which enabled robustness across varying conditions without compromising performance in ideal settings.

\heading{Transparent and Specular Surfaces} A separate challenge involves estimating depth for transparent or mirror (ToM) surfaces~\cite{Ramirez2023b,Ramirez2024,Ramirez2025}. Costanzino et al.~\cite{Costanzino2023} pioneered work dedicated to this problem, introducing specialized datasets~\cite{Ramirez2022,Ramirez2023}. Their approach leveraged segmentation maps or pre-trained networks, generating pseudo-labels by inpainting ToM objects and processing them with pre-trained depth models~\cite{Ranftl2020}, enabling existing networks to be fine-tuned for handling transparent and specular surfaces. Similarly, ~\cite{tosi2024diffusion}, instead, presented a novel approach using depth-aware diffusion models to systematically generate challenging scenes (both for adverse weather and ToM objects) with associated depth information and fine-tune networks through self-distillation.

\section{Monocular Depth Estimation Challenge} \label{sec:meth}
The fourth edition of the \acl{mdec} was hosted on CodaLab\footnote{\challink\vspace{-2pt}}~\cite{Pavao2022} as part of the CVPR'2025 workshop.
The participation instructions included a new starter pack\footnote{\codelink} supporting train-free minimal effort inference with public foundation models.

The development phase lasted four weeks, using the \acl{syns} validation split.
During this phase, the leaderboard was public, but the participants' usernames were anonymized.
Each participant could see the results achieved by their submissions.
The final phase of the challenge was open for three weeks.
At this stage, the leaderboard was completely private, disallowing participants to see their scores to avoid overfitting to the test split. 

Following the third edition \cite{Spencer_2024_CVPR}, any form of supervision was allowed; 
as long as only one input image is used to obtain the prediction corresponding to the input image. 
This measure prevents mixing in methods with strong support of geometric priors.
In what follows, we report results only for submissions that outperform the specified baselines in at least one point cloud or image-based metric in the testing phase.

\heading{Dataset}
The challenge is based on the \acl{syns} dataset~\cite{Adams2016,Spencer2022}, chosen due to the diversity of scenes and environments. 
A few representative examples are shown in \fig{data:syns}.
\acl{syns} also provides extremely high-quality dense ground-truth \acs{lidar}, with an average coverage of 78.20\% (including sky regions). 
Given such dense ground truth, depth boundaries were obtained through edge detection on the log-depth maps, allowing us to compute additional edge-based metrics.
As outlined in~\cite{Spencer2022,spencer2023second,Spencer_2024_CVPR}, the images were manually checked to remove dynamic object artifacts.

\heading{Evaluation}
Participants were asked to provide predictions for each dataset image. 
This year, we switched to LSE-based alignment between predictions and ground truth maps to accept various types of predictions. 
In addition to previously accepted \texttt{disparity} prediction methods, we welcomed \texttt{affine-invariant}, \texttt{scale-invariant}, and \texttt{metric} types.
The evaluation server bilinearly upsampled the predictions to the target resolution.
Submissions with disparity images (including previous years' submissions) were inverted into scale-invariant depth maps.
Together with the other prediction types, we performed least squares alignment with two degrees of freedom, the affine-invariant scale and shift, and computed multiple pairwise metrics.

\heading{Metrics}
Following the previous editions of the challenge~\cite{Spencer2023,spencer2023second,Spencer_2024_CVPR}, we use a combination of image-, pointcloud-, and edge-based metrics.
Image-based metrics are the most common (MAE\down, RMSE\down, AbsRel\down, $\delta$\up) and are computed using pixel-wise comparisons between the predicted and ground-truth depth map. 
Pointcloud-based metrics~\cite{Ornek2022} (F-Score\up) instead bring the evaluation into the 3D domain.
Among these, we select the reconstruction F-Score as the leaderboard ranking metric. 
Finally, edge-based metrics are computed only at depth boundary pixels. 
This includes F-Score at edges (F-Edges\up), as well as edge accuracy and completion metrics from IBims-1~\cite{Koch2018}.

\section{Challenge Submissions} \label{sec:submit}
We now report the main technical details for each submission, as provided by the participants themselves. 

\subsection*{Baseline \#1 -- Marigold \cite{ke2024repurposing}}
\noindent
Baseline model for this year's challenge.
\\
\heading{Network}
Based on the convolutional UNet LDM~\cite{rombach2022high}, repurposed for depth estimation. The model gradually denoises a random latent code conditioned on the input image to produce a depth map. It uses the frozen VAE component from the LDM for encoding and decoding, with the depth map replicated into three channels to match the RGB input format. The denoising UNet is modified to accept both image and depth latent codes concatenated along the feature dimension.
\\
\heading{Supervision}
The model is fine-tuned from the LDM~\cite{rombach2022high} using only synthetic datasets (Hypersim~\cite{Roberts2021}, Virtual KITTI 2~\cite{Cabon2020}) to ensure dense, complete, and clean depth maps. It uses affine-invariant depth normalization through a linear transformation that maps depth values to the range $[-1, 1]$.
\\
\heading{Training}
Trained for 18K iterations with batch size 32 (accumulated over 16 steps to fit on a single GPU) using Adam optimizer with a learning rate of $3e^{-5}$ and random horizontal flipping augmentation. The training takes approximately 2.5 days on a single consumer GPU. 
\\
\heading{Inference}
Off-the-shelf \texttt{prs-eth/marigold-depth-v1-0} checkpoint is applied at the native input resolution with 10 DDIM steps and ensemble size 1\footnote{\label{fn:reproducibility}\href{https://github.com/toshas/mdec_benchmark/blob/main/mdec_2025}{https://github.com/toshas/mdec\_benchmark/blob/main/mdec\_2025}}.

\subsection*{Baseline \#2 -- Depth Anything v2 }
\noindent
Additional baseline model for this year's challenge.
\\
\heading{Network}
DPT decoder \cite{Ranftl2021} built on DINOv2 encoders \cite{oquab2023dinov2}. Multiple variants are available, from lightweight (ViT-S, 25M parameters) to high-capacity (ViT-G, 1.3B parameters).
\\
\heading{Supervision}
Three-stage training pipeline: 1) Train a teacher model (DINOv2-G) purely with high-quality synthetic images from five datasets (595K images); 2) Produce precise pseudo-labels on large-scale unlabeled real images dataset; 3) Train a final student model on 62M pseudo-labeled images from eight datasets. Training losses include a scale- and shift-invariant loss, gradient matching, and feature alignment to preserve semantics from pre-trained DINOv2 encoders.
\\
\heading{Training}
Trained at resolution $518 \times 518$ by resizing input to match the shorter side, followed by random cropping. The teacher model is trained with batch size 64 for 160K iterations. Student models are trained with batch size 192 for 480K iterations. Adam optimizer is used with learning rates $5e^{-6}$ (encoder) and $5e^{-5}$ (decoder). %
\\
\heading{Inference}
Off-the-shelf \texttt{depth-anything/Depth-Anything-} \texttt{V2-Large-hf} checkpoint is applied at native resolution\footref{fn:reproducibility}.

\subsection*{\teamsec{1}}
\emph{%
\begin{tabular}{ll}
            S.\ Wang & wangshuaihang@hikvision.com \\
            Z.\ Ma & mazhenxin@hikvision.com \\
            W.\ Chen & chenweijie5@hikvision.com \\
            B.\ Xu & xubaobei@hikvision.com \\
            F.\ Sun & sunfengyu@hikvision.com \\
            D.\ Xie & xiedi@hikvision.com \\
            J.\ Zhu & zhujiang.hri@hikvision.com \\
\end{tabular}
}

\noindent
\heading{Network}
ViT-G encoder with DPT decoder head. The encoder is initialized with pre-trained weights from Metric3D v2 \cite{hu2024metric3d}, while the decoder uses random initialization.\\
\heading{Supervision}
Supervised learning with SILog, SSIL, and Feature Alignment losses. Feature Alignment uses KL divergence between the network encoder output and frozen pre-trained DINOv2 (ViT-G) encoder features for semantic knowledge distillation. The upper quarter and lateral sixteenths of label maps were excluded from loss computation due to low-quality annotations.\\
\heading{Training}
Cityscapes dataset with pixel-wise depth ground truth. Learning rates $1e^{-5}$ for encoder and $1e^{-4}$ for decoder, batch size 32 across 8 GPUs, with random horizontal flipping. Early stopping at 3 epochs to mitigate overfitting. Training images were resized to $384 \times 768$, testing images to 768px on the short side maintaining aspect ratio.

\subsection*{\teamsec{2}}
\emph{%
\begin{tabular}{ll}
            M.\ Lavreniuk & nick\_93@ukr.net \\
\end{tabular}
}

\noindent
\heading{Network}
Based on Depth Anything v2 \cite{Yang2024v2}. 
\\
\heading{Supervision}
Supervised training with ground-truth depth using scale-invariant logarithmic loss, scale-and-shift invariant loss, and random proposal normalization loss.
\\
\heading{Training}
Trained on outdoor datasets (KITTI \cite{Geiger2013}, Virtual KITTI 2 \cite{Cabon2020}, DIODE outdoor \cite{diode_dataset}, Cityscapes \cite{Cordts2016}) with random crop size \shape{518}{1078}{}{}, batch size 1, learning rate $1e^{-6}$, maximum depth threshold 80 meters, for 5 epochs. Standard augmentations were applied during training and avoided during testing due to performance degradation.

\subsection*{\teamsec{3}}
\emph{%
\begin{tabular}{ll}
            H.\ Guan & guanhaining@megvii.com \\
            Q.\ Wu & wuqun@megvii.com \\
            Y.\ Zeng & zengyupei@megvii.com \\
            C.\ Lu & luchao03@megvii.com \\
            H.\ Wang & wanghuanran@megvii.com \\
\end{tabular}
}

\noindent
\heading{Network}
Depth-Anything-V2 (ViT-L) \cite{Yang2024v2}, using the pre-trained outdoor metric depth estimation model.
\\
\heading{Supervision}
Fine-tuned on ground truth metric depths using a composite loss comprising SiLog, SSIL, Virtual Normal, and Sky Regularization.
\\
\heading{Training}
Fine-tuned on Cityscapes \cite{Cordts2016}, KITTI \cite{Geiger2013}, and VKITTI2 \cite{Cabon2020} with $392 \times 770$ resolution, random horizontal flip, and canonical space camera transformation. Depth prediction up to 120m with dataset-specific truncation. Sky regions forced to maximum depth. Trained for 5 epochs with batch size 8, learning rates $5e^{-6}$ for encoder and $5e^{-5}$ for decoder. Inference uses test-time resolution scaling to 518px on the shorter side.

\subsection*{\teamsec{4}}
\emph{%
\begin{tabular}{ll}
G.\ Zhou & zhouguangyuan@buaa.edu.cn \\
H.\ Zhang & zht199599@gmail.com \\
J.\ Wang & doublewjx@gmail.com \\
Q.\ Rao & erdosv001@gmail.com \\
C.\ Wang & chunjiewang1993@gmail.com \\
X.\ Liu & liuxiao@foxmail.com \\
\end{tabular}
}

\noindent
\heading{Network}
Based on Depth-Anything \cite{Yang2024} with a BEiT384-L backbone.
\\
\heading{Supervision}
Supervised using stereo disparities from Cityscapes \cite{Cordts2016}. The final loss is composed of the SSIL loss, Gradient Angle loss, and random proposal normalization loss (RPNL) loss.
\\
\heading{Training}
The network was fine-tuned at a resolution of $384 \times 768$ with the ground truth depth resolution of $1024 \times 2048$. Only random flipping was used as the data augmentation strategy. The training was restricted to 4 epochs, a strategic choice to prevent overfitting and ensure the model's robustness to new data.

\subsection*{\teamsec{5}}
\emph{%
\begin{tabular}{ll}
            Z.\ Lou & johnlou@insta360.com \\
            H.\ Jiang & jianghualie@insta360.com \\
            Y.\ Chen & chenyihao@insta360.com \\
            R.\ Xu & jerry1@insta360.com \\
            M.\ Tan & tmltan@insta360.com \\
\end{tabular}
}

\noindent
\heading{Network}
Pre-trained ViT-Large variant of Depth-Anything-V2 \cite{Yang2024v2} with the final ReLU layer replaced by Softplus activation for more stable training. Final output converted to affine-invariant depth values using shifted inverse function: $\text{Depth} = \frac{200}{1+\text{DaV2}(\text{img})}$.
\\
\heading{Supervision}
Uses scale-invariant log loss, scale-shift-invariant loss and multi-scale gradient matching loss. For synthetic datasets (Virtual KITTI 2 \cite{Cabon2020}, Hypersim \cite{Roberts2021}), official ground truth depth is used. For real-world datasets (CityScapes, KITTI), dense and detailed depth maps are generated using FoundationStereo and DeformStereo.
\\
\heading{Training}
Trained on CityScapes, KITTI, VKITTI2, and Hypersim, preprocessed to match SYNS-Patches vertical-to-horizontal FOV ratio. Input size $392 \times 784$, batch size 8, with learning rates $1.5e^{-5}$ for encoder and $1e^{-4}$ for decoder, fine-tuned for 5 epochs.

\subsection*{\teamsec{6}}
\emph{%
\begin{tabular}{ll}
            Z.\ Qin & 24s103315@stu.hit.edu.cn \\
            Y.\ Mao & 24s103391@stu.hit.edu.cn \\
            J.\ Liu & 1190200503@stu.hit.edu.cn \\
            J.\ Xu & xujialei@stu.hit.edu.cn \\
            Y.\ Yang & 2022111217@stu.hit.edu.cn \\
            W.\ Zhao & wbzhao@hit.edu.cn \\
            J.\ Jiang & jiangjunjun@hit.edu.cn \\
            X.\ Liu & csxm@hit.edu.cn \\
\end{tabular}
}

\noindent
\heading{Network}
Based on Depth Anything v2 \cite{Yang2024v2} with a ViT-L backbone, initialized with the authors' pre-trained weights for indoor metric depth. The entire pre-trained backbone is frozen to preserve its generalization capability while a parallel trainable branch is added. This trainable branch is a full copy of the original model, connected via zero-initialized $1 \times 1$ convolutions to the frozen backbone, outputting per-pixel scale ($\alpha$) and shift ($\beta$) maps to refine the base prediction through $D_{\text{final}} = (1 + \alpha) \odot D_{\text{base}} + \beta$.
\\
\heading{Supervision}
Fine-tuned in a supervised manner using the SSIL loss as the loss function.
\\
\heading{Training}
Fine-tuned on CityScapes \cite{Cordts2016} and DIODE \cite{diode_dataset} datasets with inputs randomly cropped to $768 \times 768$ and augmented via random horizontal flipping. Trained with batch size 32 and AdamW optimizer with weight decay 0.01. Learning rates set to $5e^{-6}$ for pre-trained parameters and $5e^{-5}$ for newly introduced ones. OneCycleLR schedule adopted, starting at $1.25e^{-6}$ (one-fourth of base rate), increasing to $5e^{-6}$ after 20\% warmup phase, and subsequently decaying to $2.5e^{-6}$.

{ %
\addtbl*[!t]{results}{res:results}
}

\subsection*{\teamsec{7}}
\emph{%
\begin{tabular}{ll}
            M.\ Zhao & mingshuai\_z@bupt.edu.cn \\
            A.\ Ming & mal@bupt.edu.cn \\
            W.\ Chen & chenw@bupt.edu.cn \\
            F.\ Xue & feng.xue@unitn.it \\
            M.\ Yu & yumengying@bupt.edu.cn \\
            S.\ Gao & gaostar2024@bupt.edu.cn \\
            X.\ Wang & xiangfeng\_w@foxmail.com \\
\end{tabular}
}

\noindent
\heading{Network}
Improved SM4Depth \cite{liu2024sm4depth} based on Booster \cite{Ramirez2022}. The architecture includes FOV alignment preprocessing, an encoder from Depth Anything \cite{Yang2024}, pyramid scene transformer (PST) with three parallel transformer encoders, online domain-aware bin mechanism, and a decoder with hierarchical scale constraints (HSC-Decoder). The encoder is frozen while only the PST and the Decoder are fine-tuned.
\\
\heading{Supervision}
Multiple loss functions: SILog loss, gradient matching loss, pixel-wise loss, virtual normal loss, bi-directional Chamfer loss, and cross entropy loss combined as $L = L_{pixel} + \sum_{s=2}^4 L_{pixel_s} + \mu L_{vnl} + \gamma L_{bin} + L_{cel}$ where $\mu=5$ and $\gamma=0.1$.
\\
\heading{Training}
DinoV2 \cite{oquab2023dinov2} Base backbone, trained for 20 epochs with batch size 10 on a single consumer GPU. Adam optimizer with $(\beta_1, \beta_2) = (0.9, 0.999)$, initial learning rate $2e^{-5}$ gradually reduced to $2e^{-6}$. Fixed FOV set to $(\omega_x^{\prime}, \omega_y^{\prime}) = (58^\circ, 45^\circ)$ and fixed resolution to $564 \times 424$.

\subsection*{\teamsec{8}}
\emph{%
\begin{tabular}{ll}
            G.\ Omotara & goowfd@missouri.edu \\
            R.\ Farag & rmf3mc@missouri.edu \\
            J.\ Demby's & udembys@missouri.edu \\
            S.\ M.\ A.\ Tousi & stousi@missouri.edu \\
            G.\ N.\ DeSouza & desouzag@missouri.edu \\
\end{tabular}
}

\noindent
\heading{Network}
Builds on Marigold~\cite{ke2024repurposing} by integrating semantic information. While the Marigold network remains frozen, a semi-supervised model (UniMatch-V2 \cite{yang2025unimatch}) generates semantic segmentation masks that capture rich scene semantics. A new DenseNet-like \cite{huang2017densely} model is trained to incorporate the extracted semantics to refine and enhance the depth map.
\\
\heading{Supervision}
Fully supervised training using SILog loss to optimize the depth predictions.
\\
\heading{Training}
Trained on DIODE \cite{diode_dataset} dataset. Ground truth depth maps are first scaled, shifted, and normalized to $[0, 1]$ to match Marigold's affine-invariant representation. Training was conducted for 50 epochs with learning rate $3e^{-5}$ to ensure a smooth loss trajectory.

\subsection*{\teamsec{9}}
\emph{%
\begin{tabular}{ll}
            T.\ A.\ Yang & ytanh21@apcs.fitus.edu.vn \\
            M.\ Q.\ Nguyen & nmquang21@apcs.fitus.edu.vn \\
            T.\ P.\ Tran & ttphuc21@apcs.fitus.edu.vn \\
\end{tabular}
}

\noindent
\heading{Network}
Hybrid approach combining Felzenszwalb segmentation and segment-wise confidence heuristic to improve depth accuracy. Segments images into consistent regions, facilitating depth estimation within structurally coherent areas. Combines outputs from diffusion-based (Marigold \cite{ke2024repurposing}) and transformer-based (Depth Anything v2 \cite{Yang2024v2}) models.
\\
\heading{Supervision}
Segment-wise confidence computation using inverse depth variance to determine the reliability of depth estimates. Confidence metric defined as $c_i = \frac{1}{\sigma_i + \epsilon}$ where $\sigma_i$ is the depth variance within segment $i$. Adaptive depth selection enables the fusion of different model outputs at the segment level based on the confidence threshold.
\\
\heading{Training}
Felzenszwalb segmentation parameters: scale=200, $\sigma$=0.8, min\_size=50. Fusion strategy uses confidence threshold T=1.2, preferring Marigold predictions unless Depth Anything v2's confidence is significantly higher.

\subsection*{\teamsec{10}}
\emph{%
\begin{tabular}{ll}
            A.\ Luginov & a.luginov@pgr.reading.ac.uk \\
            M.\ Shahzad & m.shahzad2@reading.ac.uk \\
\end{tabular}
}

\noindent
\heading{Network}
Custom hybrid CNN-Transformer architecture with SwiftFormer-L1 \cite{shaker2023swiftformer} encoder and a two-level decoder similar to SwiftDepth \cite{luginov2023swiftdepth}. A separate ResNet-18 camera network predicts pose and intrinsics during training only. Both encoders are pre-trained on ImageNet \cite{deng2009imagenet}. The depth network has 30M parameters, while the camera network has 15M parameters.
\\
\heading{Supervision}
Combined self-supervised learning (SSL) and pseudo-supervised learning (PSL) \cite{luginov2024nimbled}. SSL uses minimum reprojection loss with auto-masking, PSL uses scale-and-shift invariant loss, with Depth-Anything-V2-Large generating pseudo-labels.
\\
\heading{Training}
Trained simultaneously on YouTube \cite{luginov2024nimbled} (1.1M triplets) and KITTI \cite{Geiger2013} (40K triplets). Batch size 16, learning rate $1e^{-5}$, weight decay $1e^{-3}$, for 110 epochs on KITTI and 4 epochs on YouTube. The loss weighting parameter was adjusted from 0.9 to 0.95 during training.

\section{Results} \label{sec:res}

Following the protocol from previous editions, submitted predictions were evaluated on the testing split of \acl{syns}~\cite{Adams2016,Spencer2022}, 
after being aligned to ground-truth depths according to median depth scaling or least-squares alignment, as requested by the participants.

\addfig*[t]{depth}{res:depth_viz}

\subsection{{Quantitative Results}}

\tbl{res:results} collects the results of this fourth edition of the challenge, ranking the submitted methods according to their F-Score performance. Within the table, we highlighted the new \colorbox{Dsectioncolor}{baselines} introduced for this edition, as well as the results achieved by the \colorbox{Asectioncolor}{winning team of the third edition} according to the new aligning protocol -- both excluded from the official ranking. We also grouped entries with close-to-identical results within brackets, likely representing duplicate team submissions.

Similarly to the previous edition \cite{Spencer_2024_CVPR}, where the Depth Anything model \cite{Yang2024} was widely adopted by participants, this year we observe a widespread use of its successor, Depth Anything v2 \cite{Yang2024v2}, among the leading teams -- including \team{2}, \team{3}, \team{5}, \team{6}.

Team~\team{1} secured the top position on the leaderboard with an F-score of 23.05, surpassing the best baseline method (Marigold) by about $35\%$. Among the teams submitting the report, \team{1} is the only one outperforming PICO-MR, the winning team from ``The Third Monocular Depth Estimation Challenge'' \cite{Spencer_2024_CVPR}, in terms of F-score.

Team \team{2}, in the third position, achieved an F-score lower than what was achieved by PICO-MR (20.81) while outperforming it on several image-based metrics. Peculiarly, \team{2} achieves the absolute best results on AbsRel and $\delta<1.25^3$ metrics. 

Teams \team{3} and \team{4}, ranking fourth and sixth, respectively, obtained results very close to those of \team{2}, although not excelling in any particular metric. Nonetheless, they could still outperform Marigold by approximately 20\% on the F-score. 
In contrast, team \team{5} performed slightly worse in terms of F-score while securing the absolute best results on metrics such as Acc-Edges, $\delta<1.25$ and $\delta<1.25<2$. Similarly, team \team{6} achieved the absolute best result for F-Edges. In contrast, team \team{7} did not excel in most metrics while consistently outperforming the baselines in most metrics, such as F-score, MAE, RMSE, and AbsRel.

Finally, teams \team{8}, \team{9}, and \team{10} managed to outperform the baselines on only one metric among the ten evaluated. We highlight the efforts by these teams to develop custom solutions not limited to fine-tuning existing foundational models. Their performance further underscores the importance of the web-scale data used to train such foundational models, giving them a clear advantage over any possible alternative that does not leverage similar quantities of training data.

\subsection{Qualitative Results}

\fig{res:depth_viz} presents selected qualitative results from the SYNS-patches test set. From top to bottom, we report input images and ground truth depth maps, followed by the predictions from teams that submitted the report, ordered according to their achieved F-score. For reference, we also show qualitative results from PICO-MR \cite{Spencer_2024_CVPR} -- characterized by noticeable grid artifacts. 

We appreciate how most top-performing methods retain significantly more detailed predictions than PICO-MR. This enhanced level of detail is caused by the use of more modern foundational models, such as Metric3D v2 \cite{hu2024metric3d} and Depth Anything v2 \cite{Yang2024v2}, which inherently generate more refined depth maps than the original Depth Anything.
Furthermore, we appreciate how the predictions by \team{1} and particularly \team{2} appear cleaner and exhibit fewer artifacts, pointing at a general qualitative improvement beyond the quantitative ones observed in the tables.

This widespread adoption of depth foundation models led to substantial improvements in both indoor and outdoor scenarios, proving once again the crucial role of web-scale data availability, further pushing the boundaries of MDE.
However, similarly to previous editions, all methods still exhibit over-smoothing issues at depth discontinuities, which can be seen as content bleeding artefacts when depth maps are projected into point clouds. 

To summarize, the qualitative results provide additional insights beyond those that emerged from the quantitative analysis, hinting at further improvements over the previous editions. 
Nonetheless, evident challenges persist when aiming at generalization, such as dealing with very thin structures or non-Lambertian surfaces, leaving room for further developments in the field.

\section{Conclusions \& Future Directions} \label{sec:conclusion}
This paper summarized the results of the fourth edition of \ac{mdec}.
Compared to previous editions, we have seen a further increase in participation in our challenge, confirming that MDE is a vibrant research trend. We witnessed a drastic increase in reliance on the pre-trained foundational models and various prediction types. 

Despite yielding marginal improvements over the previous years in terms of the F-score, this year's methods generally deliver sharper predictions and appear useful in downstream tasks. 
This suggests that, despite the significant improvements achieved by foundational models over previous approaches, the problem remains far from being solved.
As the benchmark approaches saturation, further progress will likely require more than simply increasing the volume of training data or usage of foundation model priors.
Next-generation advances in depth estimation may come from tackling more challenging settings, such as the perception of non-Lambertian surfaces \cite{Ramirez2022}, accurate metric depth estimation \cite{hu2024metric3d}, or alternative scene representations, such as point maps~\cite{wang2024dust3r}.

We hope future editions of \ac{mdec} will attract even more participants to push the boundaries of \ac{mde} forward.

{
\small
\bibliographystyle{ieee_fullname}
\bibliography{main}
}

\end{document}